\def\year{2022}\relax

\documentclass[letterpaper]{article} 

\usepackage{float}
\usepackage{multirow}
\usepackage{subfig}
\usepackage{aaai22}  
\usepackage{times}  
\usepackage{helvet}  
\usepackage{courier}  
\usepackage[hyphens]{url}  
\usepackage{graphicx} 
\usepackage{natbib}  
\usepackage{caption} 
\DeclareCaptionStyle{ruled}{labelfont=normalfont,labelsep=colon,strut=off} 
\usepackage{xcolor}
\usepackage{algorithm}
\usepackage{algorithmic}

\usepackage{newfloat}
\usepackage{listings}
\floatstyle{ruled}
\newfloat{listing}{tb}{lst}{}
\floatname{listing}{Listing}

\title{
Toxicity Inspector: A Framework to Evaluate Ground Truth in Toxicity Detection Through Feedback
}
\author{
   Huriyyah Althunayan, Rahaf Bahlas, Manar Alharbi, Lena Alsuwailem, Abeer Aldayel, Rehab ALahmadi \\
}
\affiliations{
    King Saud University, 
College of Computer and Information Sciences\\

    Information Technology Department\\
    441201098@student.ksu.edu.sa, 441201261@ student.ksu.edu.sa, 441201217@student.ksu.edu.sa, 441201025@student.ksu.edu.sa, aabeer@ksu.edu.sa, ralahmadi@ksu.edu.sa  
   
}
\begin{document}

\maketitle

\begin{abstract}

Toxic language is difficult to define, as it is not monolithic and has many variations in perceptions of toxicity. This challenge of detecting toxic language is increased by the highly contextual and subjectivity of its interpretation, which can degrade the reliability of datasets and negatively affect detection model performance. \textcolor{black}{To fill this void, this paper introduces a toxicity inspector framework that incorporates a human-in-the-loop pipeline with the aim of enhancing the reliability of toxicity benchmark datasets by centering the evaluator's attention through an iterative feedback cycle}. The centerpiece of this framework is the iterative feedback process, which is guided by two metric types (hard and soft) that provide evaluators and dataset creators with insightful examination to balance the tradeoff between performance gains and toxicity avoidance.  
\end{abstract}

\section{Introduction}
Determining what is toxic or harmful in a language is a very subjective task, as perceived toxicity varies based on many different characteristics ~\citep{Davidson2017ai}. 

To detect toxic language, the construction of the training dataset plays an essential role in the robustness of model evaluation and performance.  
\textcolor{black}{Notably, the appropriate handling of label annotator variations during dataset construction is often overlooked. However, recent research has been conducted on alleviating the possible effects of these variations. For example, ~\citet{Waseem2017-vg} proposed a typology that differentiates between direct/indirect abusive language toward specific individuals or groups.}  Another work by ~\citet{Fanton2021-bc} proposed a human-in-the-loop revision cycle for pairs of narratives (i.e., hate speech and counter-hate speech), in which feedback is provided for the given text to generate a new dataset.

In the novel framework proposed in this study, the feedback pertains to the label, which is set as concept-shifting model.  Specifically, the feedback process is guided using soft and hard metrics, where cross entropy (CE) and error rate are provided alongside hard metric evaluations. 

Current toxicity detectors' reliability and robustness suffer from high subjectivity and bias towards keywords ~\citep{Sap2022-lj}. In this demonstration paper, a means of evaluating reliability while improving toxicity labeling is provided by incorporating a human-in-the-loop pipeline to evaluate concept shifts within a dataset.  

\textcolor{black}{Given the multitude of terms and definitions related to hate speech in the literature, several recent studies have investigated their common aspects in terms of language detection tasks. Specifically for hate speech, the research of ~\citep{Fanton2021-bc} focused on covariate shifts (modified text) in the human-in-the-loop pipeline, demonstrating that assessing covariate shifting is beneficial to addressing the shortcomings of extant collection strategies that grant either quality or quantity, but not both ~\citep{Bhatt2021-em}. This trend of diligence is clear evidence of the urgency of finding sustainable and data-centric ways to support a full benchmark creation cycle for toxicity identification in a way that provides clear transparency and evaluation capabilities between each feedback iteration. }

The toxicity inspector provides the means of iterating sections of the dataset to verify its labels through a series of feedback cycles.

\section{Toxicity Inspector}

The toxicity inspector \footnote{Beta version can be accessed from this link \url{http://toxicityinspect.com/}}comprises different modules that facilitate the examination of toxicity annotations in a given dataset. The centerpiece of the inspector is its feedback framework, whose intended utility is to provide a unified examination of toxicity benchmarks based on an expert human evaluator. The need to examine the benchmark dataset can be categorized as follows, 1) gold-labeled dataset verification and 2) noisy labeled dataset verification. For the gold-labeled dataset verification, the labels assessment process can be carried out as either the verification of an already constructed benchmark dataset with hard labels ~\citep{Plank2022-eo}, or to examine the outcome of merged datasets and evaluate the labels of the newly added instances. Another usage is to verify the noisy labels by enabling reannotation by using a seed of labels generated by a black-box API toxicity detection system. This section details the main functionality of the feedback framework provided by the Toxicity Inspector.

\begin{figure*}[ht!]
\begin{minipage}{\columnwidth}
  \centering
  \includegraphics[width=\columnwidth]{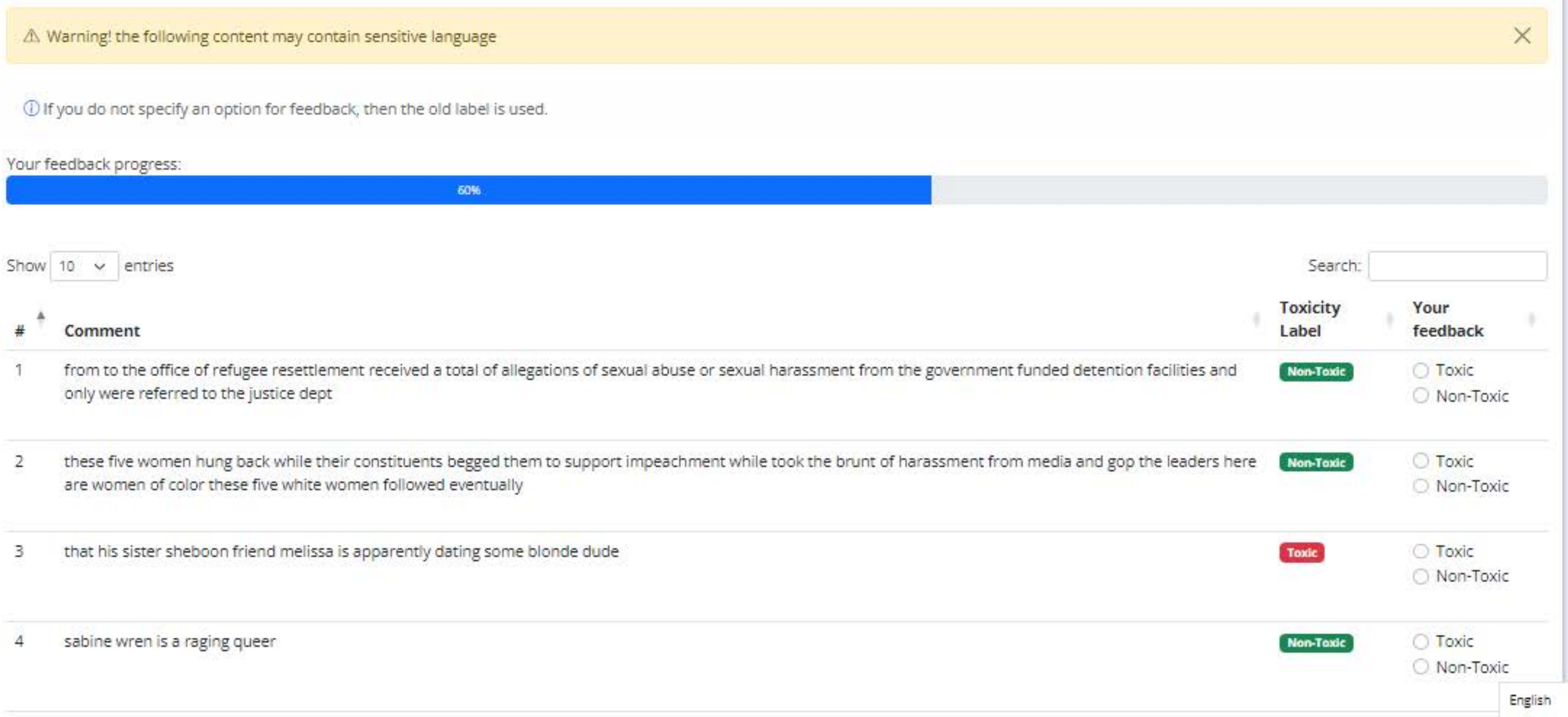}
  \captionof{figure}{Toxicity labeling feedback iterations showing the progress of corrections.}
  \label{fig:feedback}
\end{minipage}%
\hfill 
\begin{minipage}{\columnwidth}
  \centering
  \includegraphics[width=\columnwidth]{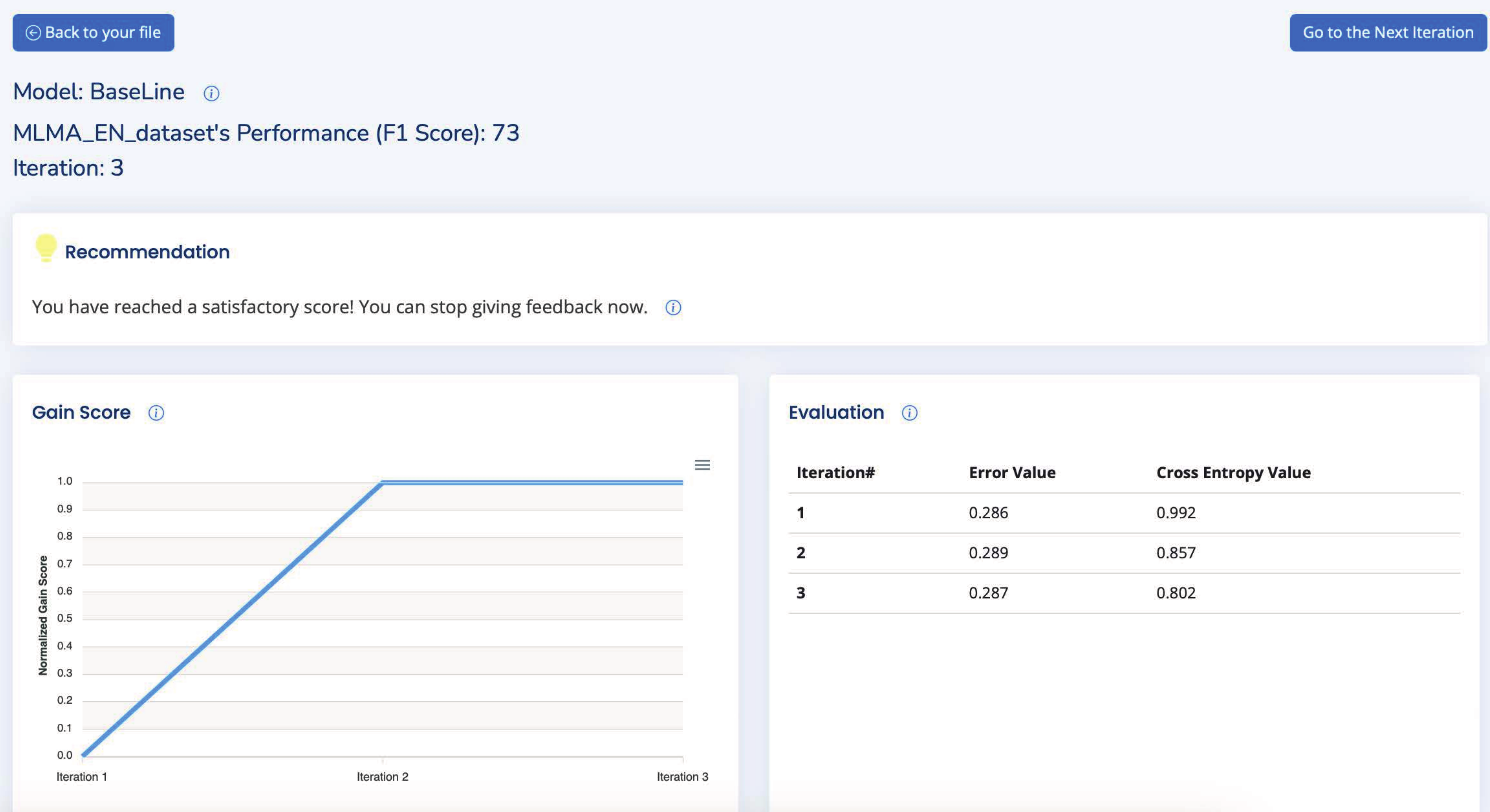}
  \captionof{figure}{Evaluation cycle using soft and hard metrics to review the full performance of each iteration. Mean squared error (MSE), cross entropy (CE), and normalized gain scores of toxicity model detection reflects the differences among model performance. The higher the gain score and lower the CE, the better the feedback iteration.}
  \label{fig:test2}
\end{minipage}
\end{figure*}

\subsection{2.1 Toxicity Diagnose}

The evaluator uploads a dataset containing text-based comments and sets the basic configuration for the task. The toxicity diagnosis function provides a preliminary overview of the overall toxicity distribution within the dataset. The evaluator then chooses between \textbf{the gold-standard evaluation,} which requires the dataset to contain human-annotated hard labels based on a majority-voted gold standard ~\citep{Plank2022-eo}, or  \textbf{the black-box evaluation,} in which the dataset lacks hard labels. Hence, the comments are preliminary labeled using the widely used Perspective API \footnote{\url{https://perspectiveapi.com/}}. \textcolor{black}{In this case, the dataset will have a seed-noisy labeling option that guides the feedback iteration through an active-learning correction process in which the input data contain noisy labels and the evaluator provides modifications ~\citep{Kremer2018-dm}. Ultimately, the evaluator corrects the labels and gains an insightful overview of the API's performance.   }

\subsection{2.2 Feedback Iteration}
A subset of comments is displayed during the iterative process using 20\% increments of the dataset \textcolor{black}{\footnote{To minimize evaluator workload, we found that using five feedback (20\%) iterations leads to the simplest comment navigation and review effort.}}. \textcolor{black}{Comments are retrieved from both training and testing dataset splits to ensure that the corrections are sufficiently reflected in the overall concept-shifted evaluation.} Figure \ref{fig:feedback} illustrates the feedback process using an input set of comments, and corrections are made to the labels. The evaluator can receive explanations by clicking on each comment, where the most influential toxic and nontoxic terms are shown using model coefficients and Shapley additive explanation  ~\citep{Lundberg2017-gn} for local and global analyses of feature (word) importance. \textcolor{black}{This process enables the human evaluator to act as an expert during the task while providing label corrections.}

\subsection{2.3 Dynamic Evaluation }
After each iteration, the evaluation page provides the evaluator with a dynamic examination of the overall reliability of each feedback iteration. This process incorporates hard metrics (e.g., F1 score) and soft metrics (e.g., CE) that show the model's ability to capture human corrections (not just top labels) ~\citep{Plank2022-eo}. Figure \ref{fig:test2} shows the evaluation page after a feedback iteration, where the normalized gain score from the previous iteration is shown as a hard metric (F1 score) between models trained on iterated sets of corrected labels. To evaluate the soft metrics, the MSE metric is used to show the correction rate of each iteration. \textcolor{black}{Therefore, the human evaluator can gain insights into the effectiveness of the feedback as a trade-off between having high gain (F1) and low CE scores. }

\begin{table*} [ht!]
    \centering
\begin{tabular}{l|l| l | l | l   |l  | l  | l  }
 \hline
 & \textbf{Lang}&\textbf{Original} & \textbf{Iteration 1} &  \textbf{Iteration 2} & \textbf{Iteration 3} & \textbf{Iteration 4} & \textbf{Iteration 5}\\
   &     &F1 (CE)& F1  (CE)& F1  (CE)& F1  (CE)& F1  (CE) & F1  (CE) \\ 
 \hline
 \multicolumn{8}{c}{\textbf{\textcolor{black}{Case 1: Gold Labels }}}\\
 \hline
 SVM & \multirow{2}*{En}&52  (0.35)& 63  (0.5)& 66  (0.47)& 65  (0.33)& 70  (0.26) & 71  (0.29)\\
LSTM  & &94.2  (0.9)& 94.3  (0.3)& 94.3  (0.4)& 94.4  (0.4)& 94.9  (0.8) & 95  (0.1) \\
 \hline
 SVM & \multirow{2}*{Ar}& 83 (0.5)  & 83 (0.3) & 84 (0.3) & 84 (0.27) &  84 (0.27) &  85 (0.2) \\
LSTM  & & 87 (0.06) & 87 (0.06) & 87 (0.08) & 87 (0.09) & 87 (0.02)  &  88 (0.09)
 \\ 
 \hline
\multicolumn{8}{c}{\textbf{\textcolor{black}{Case 2: Noisy labels}}}\\
 \hline 
  SVM  & \multirow{2}*{EN}&75  (0.38)& 73   (0.62)& 72 (0.52)& 70  (0.43)& 70  (0.30) & 71  (0.29)\\
LSTM  & & 41  (3.9)& 46  (0.80)& 52  (0.70)& 58  (0.70)& 72  (0.80) & 95  (0.97)
 \\ 
 \hline
 SVM & \multirow{2}*{Ar} &  58 (0.40) & 59 (0.56) & 65 (0.48) & 69 (0.47) & 76 (0.47) & 85 (0.20)  \\
LSTM  &       &63 (0.60) & 63 (0.50)  & 72 (0.80) & 80 (0.20) & 84 (0.60)   &  88 (0.09)
 \\ 
  \hline
  \end{tabular}
  \caption{Results of iterative feedback increments (20\%) on the Multilingual Multi-Aspect (MLMA) hate-speech dataset for gold-standard and black-box labels, for English (EN) and Arabic (AR) sets using the macro F1 score and cross entropy.} 
  \label{tab: 1}
  \end{table*}

\section{Case study: Gold and Noisy Labels Evaluation}
\textcolor{black}{To evaluate the effectiveness of the feedback framework in its ability to examine the reliability of a toxicity benchmark across different languages, we used the Multilingual and Multi-Aspect \citep[MLMA, ][]{Ousidhoum2019-ha} hate-speech dataset following the training/testing split proposed in a previous work}   \footnote{\url{https://github.com/HKUST-KnowComp/MLMA_hate_speech}}. \textcolor{black}{We used a subset of a dataset covering English (1.9K comments) and Arabic (1.5K comments) labeled with multiple hostility types.} To unify the evaluation process, binary labeling was used in which a non-toxic label was assigned to normal instances, and all others were labeled as ``toxic'' \footnote {The mapping of labels is further explained in \url{https://huggingface.co/datasets/nedjmaou/MLMA_hate_speech}}. 
 \textcolor{black}{For a case study, we considered two models commonly used by recent benchmarks: hatEval ~\citep{Bauwelinck2019-pk} and SemEval-2021 Toxic Spans Detection ~\citep{Pavlopoulos2021-yz}. Thus, we used linear Support Vector Machine (SVM) based on a TF-IDF representation along with LSTM as two  baseline models to evaluate the reliability of the labels using gold-labeled and noisy machine-generated labeled datasets. }
 
\textcolor{black}{To examine the annotator's workload during feedback iterations, we applied two cases each to English and Arabic languages based on the MLMA dataset. Case 1 applied gold labels, and Case 2 applied black-box labels. For all trials, the evaluator assessed the correctness of noisy labels as generated by the Perspective API on MLMA instances.} Table \ref{tab: 1} lists the F1 and CE scores for the cases across five iterations, where the number of iterations needed to correct the labels was found to less when using the gold-standard compared with the black-box. \textcolor{black}{In Case 1, a steady increase in F1 scores throughout all iterations was found in both languages. For example, in the English MLMA, a substantial gain (i.e., high F1 score, low CE) was found during the initial iterations. Using the SVM and focusing on Iteration 2, the F1 score increased by +3\%, and a relatively small CE (-0.03) was found before reaching a high gain during the final iteration (F1 = 71\%; CE - 0.29). } 
In contrast, in Case 2, a prevalent fluctuation gain was seen between iterations, and satisfactory increases in F1 scores were observed in \textcolor{black}{Iterations 4 and 5.} The average MSE for the English MLMA was about 13.8\% for the gold standard and 32\% for the black box. Similarly, the MSE for the Arabic subset was about 18\% for the gold standard and 22\% for the black box. \textcolor{black}{In case 2,} the original noisy label dataset presented a 75\% F1 score, which decreased after some iterations, reaching an optimal score of 71\%. Overall, using both hard and soft metrics helps with making trade-off judgments as the evaluator is informed about the reliability of labeling during each iteration.

\section{Conclusion}
In this demonstration paper, we introduced a human-in-the-loop framework-based toxicity inspector platform that automatically provides feedback and evaluation metrics. Researchers can gain deeper and broader understandings of the characteristics of texts by assessing their labels during the full feedback iteration process using hard and soft metrics, resulting in highly transparent overviews the most contributive terms deemed toxic or nontoxic as classified by experts. This framework offers a step forward in terms of incorporating human-in-the-loop feedback mechanisms to dynamically construct optimal benchmarks for data toxicity detection. The platform provides additional modules that enable diverse evaluations based on different languages and topical characteristics. 
\subsubsection{Acknowledgments.}
This work has been partially supported by: the Deanship of Scientific Research at King Saud University for funding this work through the Undergraduate Research Support Program, College of Computer and Information Sciences. Also, we extend our appreciation to the digital innovation award (MCIT) for the initial support.  
\bibliography{main}

\end{document}